\documentclass[
]{ceurart}

\usepackage{listings}
\usepackage{subcaption}
\begin{document}

\copyrightyear{2024}
\copyrightclause{Copyright for this paper by its authors.
  Use permitted under Creative Commons License Attribution 4.0
  International (CC BY 4.0).}

\conference{Defactify 3: Third Workshop on Multimodal Fact Checking and Hate Speech Detection, co-located with AAAI 2024.}

\title{DeHate: A Stable Diffusion-based Multimodal Approach to Mitigate Hate Speech in Images}


\author[1]{Dwip Dalal}[%
email=dwip.dalal@iitgn.ac.in
]
\fnmark[1]
\address[1]{IIT Gandhinagar, India}

\author[1]{Gautam Vashishtha}[%
email=gautam.pv@alumni.iitgn.ac.in,
]
\fnmark[1]

\author[2]{Anku Rani}[%
email=ankurani@mit.edu,
]
\address[2]{MIT Media Lab, USA}

\author[3]{Aishwarya Reganti}[%
]
\address[3]{CMU, USA}

\author[4]{Parth Patwa}[%
]
\address[4]{UCLA, USA}

\author[5]{Mohd Sarique}[%
]
\address[5]{IIIT Kalyani, India}

\author[6]{Chandan Gupta}[%
]
\address[6]{IIIT Delhi, India}

\author[7]{Keshav Nath}[%
]
\address[7]{DTU, India}

\author[8]{Viswanatha Reddy}[%
]
\address[8]{UW Madison, USA}

\author[9]{Vinija Jain}[%
]
\address[9]{Stanford University, USA}

\author[9,10]{Aman Chadha}[%
]
\address[10]{Amazon GenAI, USA}
\cormark[1]
\author[11]{Amitava Das}[%
email=amitava@mailbox.sc.edu
]
\address[11]{University of South Carolina, USA}

\author[11]{Amit Sheth}[%
]

\author[12]{Asif Ekbal}[%
]
\address[12]{IIT Patna, India}

\fntext[1]{These authors contributed equally.}
\cortext[1]{Work does not relate to position at Amazon.}

\begin{abstract}
The rise in harmful online content not only distorts public discourse but also poses significant challenges to maintaining a healthy digital environment. In response to this, we introduce a multimodal dataset uniquely crafted for identifying hate in digital content. Central to our methodology is the innovative application of watermarked, stability-enhanced, stable diffusion techniques combined with the Digital Attention Analysis Module (DAAM). This combination is instrumental in pinpointing the hateful elements within images, thereby generating detailed hate attention maps, which are used to blur these regions from the image, thereby removing the hateful sections of the image. We release this data set as a part of the dehate shared task. This paper also describes the details of the shared task. Furthermore, we present DeHater, a vision-language model designed for multimodal dehatification tasks. Our approach sets a new standard in AI-driven image hate detection given textual prompts, contributing to the development of more ethical AI applications in social media. 
\end{abstract}


\maketitle

\section{Introduction}

In recent years, the surge in AI applications across various sectors has underscored the importance of large-scale, high-quality datasets. As AI transitions from a research focus to practical deployment, the challenge lies in creating appropriate datasets and data pipelines that critically affect the model's trustworthiness \citep{Liang2022}. The rapid evolution of AI, from a conceptual stage to widespread adoption in diverse applications such as product recommendations and complex data analysis, has brought to the forefront issues related to ethics, privacy, and security.

One of the challenges for ethical AI is hateful content. While large language models (LLMs) have emerged as pivotal tools in deciphering and generating human-like text, the efficacy and ethical integrity of these models are significantly challenged by the presence of hateful content in their training datasets. The proliferation of such content not only undermines the models' utility but also raises serious ethical concerns. Addressing this, our research introduces a novel approach to de-hate tasks, particularly in a multi-modal (text + image) context.

It is important to have good quality datasets to train AI systems. Deep learning models can be data-hungry and might require high-quality data for effective functioning. The quality of data is as crucial as its quantity, and data preparation is one of the most time-consuming tasks in the AI project lifecycle \citep{datatobiz2022datasets}. Quality datasets are fundamental to building any real-world AI application, but real-world datasets are often complex, messy, and unstructured. Despite the availability of open-source datasets, finding the right balance between quantity, quality, and relevancy of data for new problem statements remains a challenge \citep{datatobiz2022datasets}.

\begin{figure}[!t]
\centering
\rule{0.98\textwidth}{1pt} \\
\colorbox{blue!30}{\parbox{0.96\textwidth}{\centering\textbf{Example}}} 
\rule{0.98\textwidth}{1pt} \\
\textbf{Text 1:} slap the bitch in the fave with them hoes \\
\rule{0.98\textwidth}{1pt} \\

\textbf{Text 2:} \textcolor{red}{Generate a high-quality real looking image of scene: slap the bitch in the fave with them hoes. Ensure the image is visually realistic, hateful and not safe for work.} \\

\rule{\textwidth}{1pt} \\
\vspace{5mm} 
\includegraphics[width=0.3\textwidth]{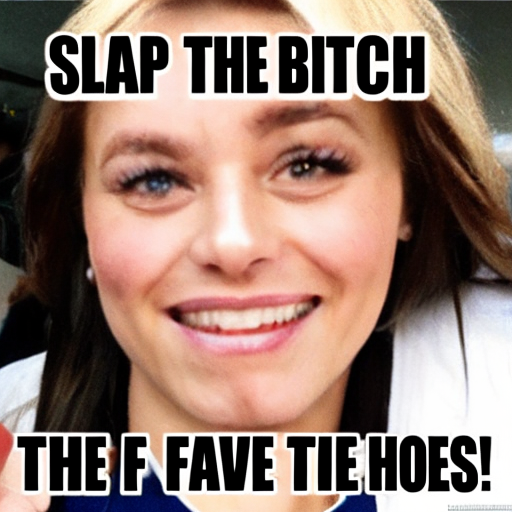} \hspace{1mm} 
\includegraphics[width=0.3\textwidth]{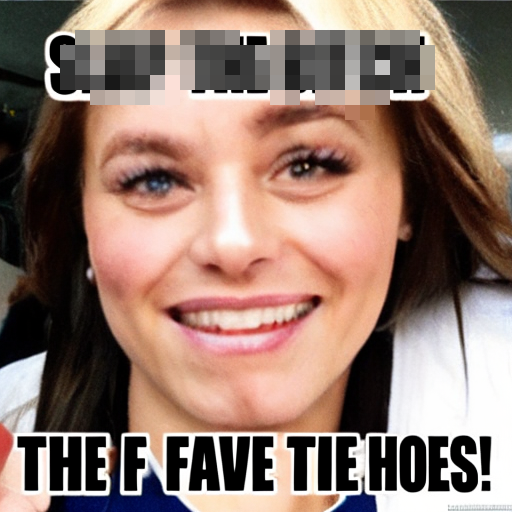} \hspace{1mm} 
\includegraphics[width=0.3\textwidth]{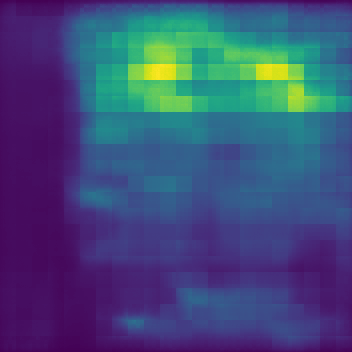} 
\rule{0.96\textwidth}{1pt} \\
\caption{A glimpse on the dataset that we curated. The first image is the image generated from the tweet, the second image is dehated version and the third image is the output of the method that we propose for dehatification task.}
\label{fig:example}
\end{figure}

Given these considerations, our research introduces a novel multimodal dataset focused on dehating digital content. Figure \ref{fig:example} provides some examples from our dataset.  Our approach includes following key contributions:

\begin{itemize}
    \item \textbf{Prompt engineering for Image-Text Alignment}: We perform prompt engineering for coherent alignment between text and images, ensuring a nuanced synchronization of multimodal content.
    \item \textbf{Multimodal Image Dehate Model Pipeline}: Our DeHater model, a unique language-image model, neutralizes hateful elements in images generated by the stable diffusion process.
    \item \textbf{Creating Multimodal DeHate Dataset}: The dataset was expanded and diversified using a Debiased LLM, developed through Dataless Model Merging, enhancing the dataset's relevance and reducing inherent biases.
\end{itemize}

In addition to creating the dataset, we also conduct 
a shared task using our data, to encourage research. The details of the dataset, our DeHate model pipeline and the shared task are described in this paper.

\section{Related Work}

In recent years, the escalation of online hate speech and hateful imagery has underscored the urgency for effective detection and mitigation strategies. Concurrently, advancements in deep learning have broadened the scope of hate detection research for english \cite{cao2020hategan,founta2018large}, other languages \cite{safi-samghabadi-etal-2020-aggression,patwa2021hater,tula2021bitions,tula2022offence,dalal2023mmt} and across modalities \cite{gunti2022memotion,yang2022multimodal,velioglu2020detecting}. \citet{masud2022proactively} meticulously compile a parallel corpus of hate texts alongside their normalized versions, offering a comprehensive view of the various contexts in which online hate speech manifests. Our research builds upon this foundation by utilizing this dataset to generate a corpus of synthetic hateful images. These images are created based on the textual content of online hate speech and are paired with regionally blurred counterparts, where the blurring is applied specifically to the areas identified as containing hate speech. This approach aligns with the growing interest in Deep Learning Model interpretability, a field that has seen rapid advancement since the advent of machine learning technologies 
\cite{chakraborty2017interpretability,huff2021interpretation,sun2021interpreting,cui2023towards,singh2024learning,rani2023sepsis,chakraborty2023factify3m,rani2023factify}.

Some researchers used cross-attention maps to enable  more profound insights into the interplay between different modalities, especially in the context of vision models \cite{tang2022daam,wu2023diffumask,dalal2023single, li2022cross}. \citet{wu2023diffumask} utilize these cross-attention insights to automatically generate accurate semantic masks and pixel-wise labels for synthetically generated images using the off-the-shelf Stable Diffusion model. Meanwhile, \cite{tang2022daam} delves into the challenges of interpretability in Latent Diffusion Models \cite{rombach2022high}, particularly in the context of attribution maps. They propose an enhanced approach by aggregating cross-attention maps in the denoising module, thus providing a more nuanced understanding of the model's decision-making process.

Our study integrates the DAAM pipeline introduced in \cite{tang2022daam} into our synthetic data generation framework. This integration facilitates the creation of hate-localized maps in images generated by the text-guided Stable Diffusion model based on the hateful subset of the MMT dataset. The DAAM pipeline is instrumental in pinpointing the hate speech span within these images, enabling us to selectively blur regions containing hateful content. This selective blurring strategy not only allows for the utilization of the non-hateful portions of the images but also contributes to the development of socially responsible AI systems.

Finally, some of the dataset on hatespeech detection include memotion datasets \cite{sharma2020semeval,ramamoorthy2022memotion,mishra2023memotion}, Multioff dataset \cite{suryawanshi2020multimodal}, OLID \cite{zampieri-etal-2019-predicting}, MMHS150K \cite{gomez2019exploringhatespeechdetection} etc while some shared tasks include \cite{,zampieri-etal-2020-semeval,patwa2021overview,patwa2022findings,thapa-etal-2023-multimodal,mishra2023memotion}.

\section{Dataset Curation}
In this section, we describe our process of dataset creation.

\subsection{Hatenorm Dataset}

The hatenorm dataset \cite{pavlopoulos-etal-2021-semeval,masud2022proactively} consists of a manually curated parallel corpus of hate texts and their normalized counterparts. The normalization process aims to make the texts less hateful and more benign. The dataset was created by sourcing hateful instances from various sources and then normalizing them to reduce the overall hatred while preserving the original semantics. Notably, the dataset captures varying degrees of hatefulness, with a focus on identifying and modifying key phrases that convey major hatred.

\subsection{DeHate Dataset}

Our dataset is crafted by leveraging the Hatenorm dataset, which is instrumental in identifying hateful content within textual data. The primary innovation in our approach lies in the utilization of the stable-diffusion-2-base model \cite{rombach2022high} for generating corresponding images. This model, known for its efficacy in image generation, was employed to transform text prompts into visual representations. The prompts were constructed by amalgamating tokens indicative of hateful terms extracted from tweets. This method ensured the generation of images that are both meaningful and contextually accurate.

However, our methodology encounters a significant constraint when dealing with extensive tweets. to solve this, we adopt a selective approach, focusing only on the most relevant segments of the text, in instances where the entire tweet exceeds the model's prompt size limit. This strategy ensures adherence to the model's technical limitations without compromising the integrity of the dataset.

The dataset comprises two distinct components:

\begin{itemize}
    \item Images Generated from Prompts: Each image in this category is a direct visual output from the stable-diffusion-2-base model, based on the processed prompt.\\
    \item Images with Blurred Hateful Components: To blur out the image components depticitng hate speech, we applied a novel technique using Diffusion Attentive Attribution Maps (DAAM) \cite{tang2022daam}. DAAM is instrumental in generating heatmaps that highlight the correlation between specific pixels and the prompt components. By computing a global heatmap value and establishing a threshold, we generated a binary mask. This mask helps identifying pixels associated with objectionable content.
\end{itemize}

\begin{figure}[!htbp]
\centering
\begin{subfigure}{.45\columnwidth}
  \centering
  \includegraphics[width=\linewidth]{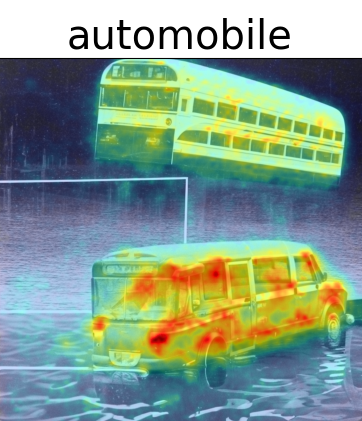}
  \caption{Automobile heat map}
  \label{fig:sub1}
\end{subfigure}%
\begin{subfigure}{.45\columnwidth}
  \centering
  \includegraphics[width=\linewidth]{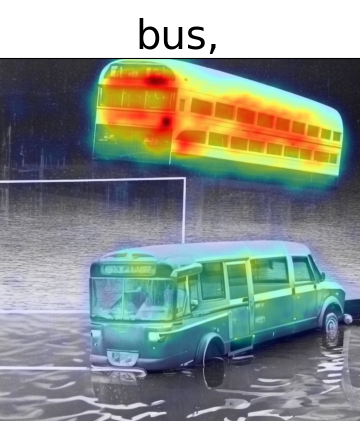}
  \caption{Bus heat map}
  \label{fig:sub2}
\end{subfigure}
\newline
\begin{subfigure}{.45\columnwidth}
  \centering
  \includegraphics[width=\linewidth]{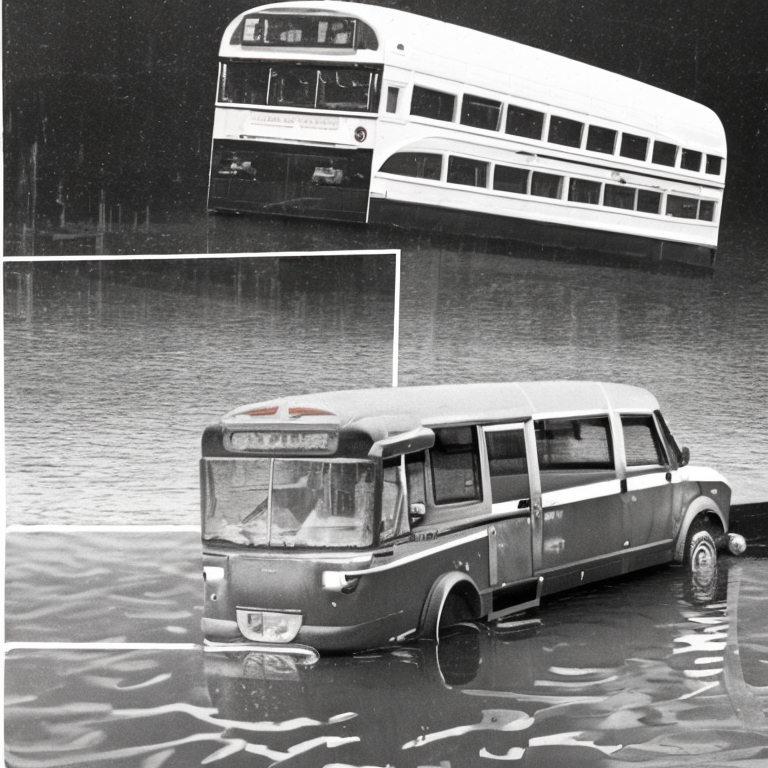}
  \caption{Output bus}
  \label{fig:sub3}
\end{subfigure}%
\begin{subfigure}{.45\columnwidth}
  \centering
  \includegraphics[width=\linewidth]{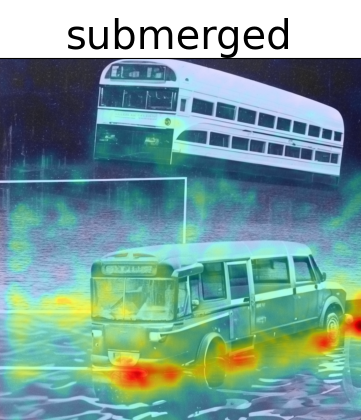}
  \caption{Submerged heat map}
  \label{fig:sub4}
\end{subfigure}
\caption{Stable diffusion output with attention maps.}
\label{fig:test}
\end{figure}

The blurring process involved two critical steps \ref{fig:test}. Initially, for each pixel in the mask with a high heatmap value, we set the corresponding pixel in a duplicate of the original image to black (RGB value: [0,0,0]), effectively 'erasing' the identified hateful element. Subsequently, for each pixel with a heatmap value of 255, we compute the average color within a localized box surrounding the pixel. This average color is then applied to the corresponding pixel in the duplicate image. This technique results in a nuanced blurring effect, where specific areas of the image are modified to represent the average color of their surroundings, thus effectively anonymizing the hateful elements while maintaining the overall context.

The final output of this process is a dataset comprising two versions of each image: the original generated version and its blurred counterpart. This dual representation serves a dual purpose: it provides a stark contrast between the unfiltered and filtered depictions of hate speech, and it offers a practical solution for training large language models in an ethically responsible manner.

The data contains total 2411 instances out of which 1687 are part of the train set and the remaining 724 are part of the test set.

\section{Shared Task details}
The dataset was released as a part of the dehate shared task at Defactify 3 workshop. Initially the labeled training data was given to the participants. later, unlabeled test set was provided. 

\subsection{Evaluation}
We use the Intersection over Union (IoU) metric to rank the participants' predictions on the test set. The IoU was computed between the predicted blurred component and the ground truth blurred component in the test dataset. 

\subsection{Participating systems}
We received 20+ registration and 5 submissions on the test set. One of the participant submitted their paper, the details are below: 

\textbf{UniteToModerate} \cite{veeramani2024unitetomoderate} use a combination of Next-Chat \cite{zhang2023nextchatlmmchatdetection} and  UniFusion \cite{qin2023unifusionunifiedmultiviewfusion} models. The NExT-Chat model provides initial mask generation through a pix2emb method, and UniFusion enhances its
precision with via hierarchical fusion of visual and reference features.

\section{Methodology}

We address the image dehatification task through an innovative approach conceptualized as unsupervised image masking. This process involves using textual prompts to guide the masking of potentially harmful areas within an image. By interpreting these prompts, the model identifies and obscures hateful content, aligning visual media with ethical standards.

Our architecture for image dehatification is based on the CLIP \cite{radford2021learningtransferablevisualmodels} (Contrastive Language–Image Pretraining) methodology, utilizing a frozen CLIP model as the encoder. This choice leverages the robust feature extraction capabilities of CLIP, which has been trained on a diverse range of images and text pairs, making it adept at understanding complex multimodal relationships.

The connection between the encoder and decoder in our model is designed by drawing inspiration from the U-Net architecture \cite{ronneberger2015unetconvolutionalnetworksbiomedical}. The U-Net-like skip connections to the CLIP encoder facilitate the transfer of rich, localized information, allowing our decoder to remain compact while retaining the essential details necessary for high-fidelity image dehatification.

Activations extracted from the encoder, including the CLS token, are then integrated into the internal activations of our decoder at each transformer block. This integration serves to enrich the decoder's understanding of the context, empowering it to generate more accurate masks for the dehatification process.

To inform the decoder about the segmentation target specifically for the task of dehatification, we employ Feature-wise Linear Modulation (FiLM) \cite{perez2017filmvisualreasoninggeneral}. FiLM allows for the modulation of the decoder’s input activation by a conditional vector that specifies the segmentation goal, enhancing the decoder's ability to focus on and accurately segment hateful content.

A pivotal part of our methodology is the use of a learnable projection network. This network combines multiple hate spans embeddings into a single projection. The employment of this learnable projection allows for the nuanced and effective condensation of varied hateful elements into a comprehensive representation, significantly improving the performance of our dehatification process.

\begin{figure}[!tbh]
\centering
\includegraphics[width=\columnwidth]{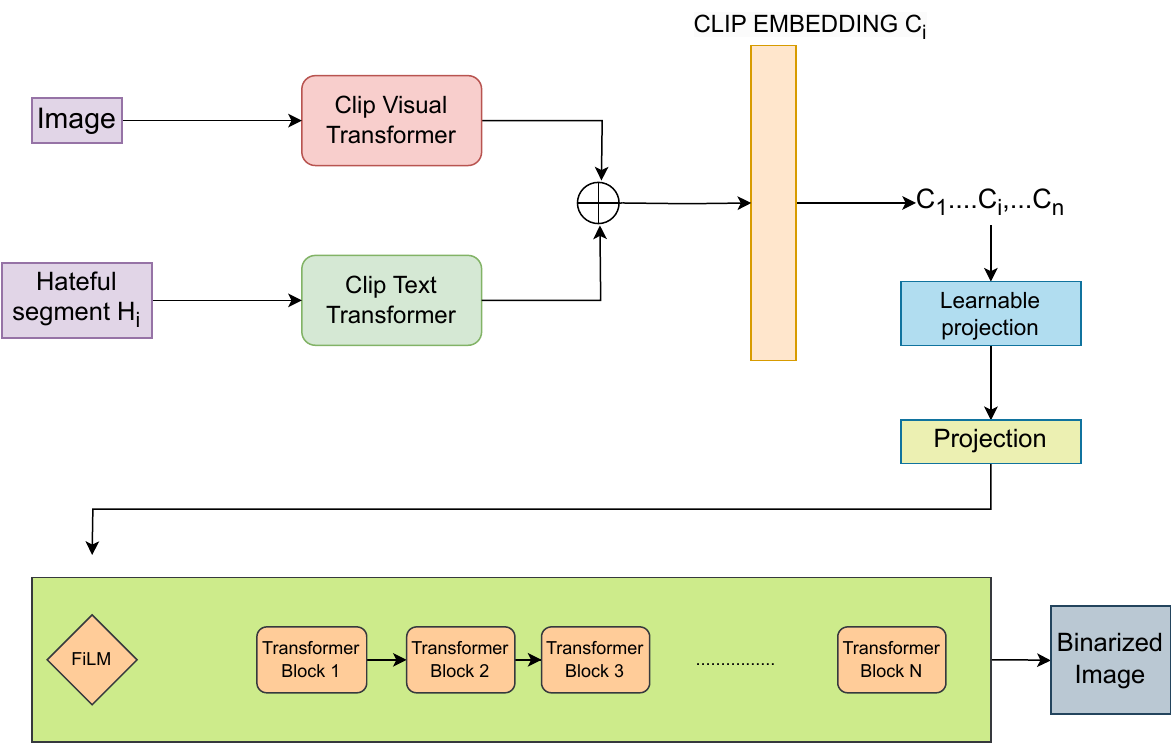}
\caption{Architecture diagram of our proposed system. }
\label{fig: parr}
\end{figure}

The output of our architecture is the production of a binarized image, which represents the final masked output. This image distinctly highlights the areas of original content deemed hateful, now masked, thus fulfilling the task of unsupervised image dehatification guided by textual prompts. An overview diagram of out method is given in figure \ref{fig: parr}.

\section{Results}

\begin{table}[]
\centering
\begin{tabular}{ccc}
%
\toprule
Rank       & Team                     & IOU score     \\ 
\midrule
\textbf{1} & \textbf{UniteToModerate} & \textbf{0.55} \\ 
2          & PaulJane                 & 0.51          \\ 
3          & Baseline (ours)          & 0.49          \\ 
4          & Markans                  & 0.48          \\ 
5          & Sanskarfc                & 0.47          \\ 
6          & rachitmodi               & 0.44          \\ 
\bottomrule
\end{tabular}
\caption{Leaderboard on the test set. }
\label{tab:leaderboard}
\end{table}

 Table \ref{tab:leaderboard} shows the official leaderboard on the test set. We can see that 2 systems outperform our baseline (0.49) whereas 3 systems do not. Team UniteToModerate \citet{veeramani2024unitetomoderate} performs the best by achieving an IOU score of 0.55. Figure \ref{fig:example} shows an example output of our method. 

\section{Conclusion and Future Work}

We introduce a dataset called deHate which pinpoints hate in multimodal content. We release the dataset as a part of a shared task to encourage research towards hate speech mitigation. Our work addresses the issue of online hate, a critical and growing concern in the digital communication landscape.

The best performing team in the shared tasks achieves an IOU score of 0.55 which shows the difficulty of the task and calls for more research. 

Future work can include using an LLM to justify the output of our hate speech mitigation pipeline. Extending our work to other languages and modalities is another direction to explore.

\bibliography{sample-ceur}


\end{document}